\documentclass{bmvc2k}

\usepackage{amsmath}
\usepackage{graphicx}
\usepackage{xcolor}
\usepackage{tikz}
\usetikzlibrary{positioning,arrows.meta,fit,backgrounds,calc}

\def\eg{\emph{e.g}\bmvaOneDot}

\def\etal{\emph{et al}\bmvaOneDot}

\title{Frugal Handwritten Text Recognition for Sub-Saharan Civil-Status Registers: A Smartphone-to-Records Pipeline with CPU-Only Inference}

\addauthor{Ousmane Hamadou}{ousmanehamadou776@gmail.com}{1}
\addauthor{Jean Louis Kedieng Ebongue Fendji}{jl.fendji@egcim-univ-ndere.cm}{2,3}

\addinstitution{
 Department of Computer Science\\
 Faculty of Science, University of Ngaound\'er\'e\\
 Ngaound\'er\'e, Cameroon
}

\addinstitution{
 Centre of Research, Experimentation and Production\\
 SCEMI, University of Ngaound\'er\'e\\
 Ngaound\'er\'e, Cameroon
}
\addinstitution{
 Stellenbosch Institute for Advanced Study\\
 Wallenberg Research Centre at Stellenbosch University\\
 Stellenbosch, South Africa
}

\runninghead{Hamadou, Fendji}{Frugal HTR for Sub-Saharan civil-status registers}

\begin{document}

\maketitle

{\renewcommand\thefootnote{}\footnotetext{\hspace{-1.9em}Accepted at the GlobalSouthCV Workshop, BMVC 2026, Lancaster, UK.}}

%=====================================================================
\begin{abstract}
Handwritten civil-status registers across Sub-Saharan Africa are being photographed, but rarely transcribed, which leaves their content unsearchable. Local administrations that hold these records typically lack GPUs, reliable connectivity, and the budget for external digitization services. We study whether a frugal pipeline, with consumer-grade capture and CPU-only inference, can make such registers machine-readable. We introduce a corpus of 1,568 smartphone images of handwritten birth, marriage, and death records from a municipal registry in Chad, with line-level transcriptions for 185 records. Our pipeline combines a fully classical preprocessing stage, running at about 2.6 images per second on a laptop CPU, with a Kraken recognizer fine-tuned from a pretrained model in about four hours on a free-tier GPU. On manually corrected line segmentation and a single-registry test set whose writers also appear in training, fine-tuning reduces the character error rate from 14.53\% to 0.50\% and the word error rate from 56.93\% to 3.37\%. On a six-core laptop CPU, the recognizer weighs 61\,MiB, peaks below 2.6\,GiB of RAM, and processes an image in under 6\,s on average, excluding line segmentation. Finally, a downstream information-extraction task shows that residual recognition errors cost about 9 F1 points on role resolution, compared with a 36.7-point gap between the fine-tuned and pretrained recognizers. These results are evidence of feasibility within one registry, not of performance on unseen writers or sites; we discuss these limitations in detail.
\end{abstract}

% =====================================================================
\section{Introduction}
\label{sec:intro}

Civil registration is the documentary basis of legal identity. Birth, marriage, and death records establish age, filiation, and nationality, and they feed the demographic statistics that public policy depends on~\cite{undesa2023,undesa2021}. In Sub-Saharan Africa, a large share of these records exists only as handwritten registers kept in municipal offices, often in poor physical condition and exposed to humidity, fire, and loss~\cite{abdoulahi2019,koaci2023}. Coverage is also low: UNICEF estimates that only about half of children under five in the region have their births registered~\cite{unicef2024}. For people without recent documents, these older registers can be the only surviving proof of identity or filiation~\cite{uneca2017}.

Several countries have launched large modernization programmes, such as Senegal's NEKKAL~\cite{nekkal2023}, which are funded at a scale most local administrations cannot match. These efforts mainly computerize new registrations or index scanned documents at the level of whole records. The historical handwritten stock is typically photographed, when it is digitized at all, but its content remains unreadable by machines: an image without a transcription cannot be searched by name, date, or place.

Handwritten text recognition (HTR) is the obvious remedy, but two obstacles stand in the way in this setting. First, \emph{domain mismatch}: publicly available HTR models are trained predominantly on European historical documents. They have never seen the handwriting styles, administrative templates, or, most importantly, the personal and place names that dominate African civil records. Second, \emph{resource constraints}: most recent HTR work assumes GPU training and inference, and often cloud processing. That is incompatible with offices that have only standard laptops, intermittent connectivity, and legitimate concerns about sending identity data to external providers.

In this paper we ask whether a frugal pipeline, meaning one captured with consumer hardware, adapted with a small amount of annotated data, and run for inference entirely on a CPU, can reach transcription quality good enough to be useful. We make the following contributions:
\begin{enumerate}
\item \textbf{A new corpus of Sub-Saharan civil-status records.} We collected 1,568 smartphone images of handwritten birth, marriage, and death records from a municipal registry in Chad, using a low-cost capture protocol. The records follow five distinct administrative templates. We provide line-level transcriptions for 185 records (Section~\ref{sec:data}).
\item \textbf{A classical, CPU-only preprocessing pipeline} for smartphone RAW captures. It handles page detection, separation of two records captured in one image, binding-margin removal, and upside-down image detection without any learned model. It ran on the full corpus without errors at about 2.6 images per second on a laptop CPU; the accuracy of its outputs was not measured (Section~\ref{sec:prepro}).
\item \textbf{An evaluation of frugal HTR adaptation.} Fine-tuning a Kraken recognizer on this small ground truth for about four hours on a free-tier GPU reduces CER from 14.53\% to 0.50\% on corrected line segmentation and writers seen in training. The resulting model runs on a laptop CPU with 61\,MiB of weights, under 2.6\,GiB of peak memory, and under 6\,s per image (Sections~\ref{sec:htr}--\ref{sec:res-frugal}).
\item \textbf{An error-propagation analysis} on a downstream information-extraction task. It quantifies how much of the end-task error comes from recognition rather than from later stages (Section~\ref{sec:res-prop}).
\end{enumerate}
Capture, preprocessing, and inference run on consumer hardware without a GPU; only fine-tuning used a free-tier cloud GPU. The stages are evaluated separately: line segmentation was corrected manually, so the pipeline is not evaluated end to end. We also report the limitations of our study openly, including the single collection site, the small test set, and the absence of inter-annotator agreement (Section~\ref{sec:disc}), and we describe how we handled the personal data involved (Section~\ref{sec:ethics}).

% =====================================================================
\section{Related Work}
\label{sec:related}

\paragraph{Handwritten text recognition.}
Handwriting recognition has long been harder than printed OCR because cursive letters join, shapes vary across writers, and character boundaries are ambiguous. This ``segmentation problem'' was the main obstacle for offline recognition~\cite{plamondon2000}. The field moved toward \emph{segmentation-free} recognition of whole text lines: recurrent networks trained with Connectionist Temporal Classification learn the alignment between image positions and output characters implicitly~\cite{graves2006ctc,graves2008mdrnn}. Convolutional-recurrent networks with CTC (CRNN-CTC) remain a standard, lightweight baseline. Kraken~\cite{kiessling2019kraken} implements this design for historical documents and, together with the eScriptorium platform~\cite{kiessling2019escriptorium}, supports the full workflow from segmentation to transcription and training. More recently, encoder-decoder Transformers such as TrOCR~\cite{li2023trocr} have achieved strong results by combining a vision Transformer encoder with a pretrained language-model decoder, at a substantially higher computational cost. We use CRNN-CTC because its cost fits our CPU inference setting. Comparing it under the same resource constraints with a Transformer recognizer, with Kraken models already trained on French handwriting (\eg from the HTR-United or CATMuS collections), with Transkribus, and with compact vision-language models run on a CPU without training remains future work.

\paragraph{Historical records and domain adaptation.}
Much HTR research on historical material relies on curated ground truth from European archives~\cite{springmann2018}, typically produced by a limited number of writers. Closest to our setting is work on demographic records: the ESPOSALLES database of historical marriage licences~\cite{romero2013esposalles} and the associated ICDAR~2017 competition on information extraction from historical handwritten records~\cite{fornes2017iehhr}. Both address recognition \emph{and} the extraction of names and roles from structured civil or parish records, which is essentially the task we study downstream. The French civil registers of births of Belfort (1807--1919) also combine pre-printed forms with handwritten fields~\cite{alkendi2024survey}. On Dutch Cura\c{c}ao death certificates (1879--1949) transcribed with a custom-trained Transkribus model, Hoek \etal found that names and professions still contain errors when the rest of the transcription appears flawless, and that whole-text quality indicators say little about the entities scholars need~\cite{hoek2026haystack}. To our knowledge, comparable resources for Sub-Saharan African civil registers do not exist. Our registers differ from European corpora in three ways: many officers wrote them with no common calligraphic standard, they mix pre-printed forms with handwritten fields, and they are dominated by local personal and place names.

\paragraph{Name bias in pretrained models.}
Pretrained language models such as CamemBERT~\cite{martin2020camembert} are trained on web text in which Sub-Saharan African names are likely under-represented. Subword tokenization splits rare names into many fragments, which weakens their representation~\cite{bostrom2020,ahia2023}. Multilingual NER performance also drops sharply for under-represented African languages~\cite{adelani2022masakhaner2}. Our setting differs: the text is in French, and it is the \emph{names} rather than the language that are out of distribution. The same concern plausibly applies to recognition models trained mostly on European material; we measure the effect of adaptation on the extraction model per entity type, but not on name spans in the recognizer.

\paragraph{Efficient deployment.}
The ``Green AI'' position argues that model quality should always be reported alongside computational cost~\cite{schwartz2020greenai}. This matters especially where access to compute and connectivity is limited, as in much of Sub-Saharan Africa, where cloud-based AI risks widening existing digital divides~\cite{idrc2026,fendji2024leftout}. Compression techniques such as post-training quantization~\cite{jacob2018quant,nagel2021whitepaper}, pruning~\cite{han2015pruning}, and distillation~\cite{sanh2019distilbert} are complementary and have been applied, for example, to deep-learning models for agricultural plants~\cite{fountsop2020compression}; their speed-ups, however, depend on hardware and implementation. Parameter-efficient fine-tuning~\cite{hu2022lora,dettmers2023qlora} reduces training memory, yet QLoRA's 4-bit kernels typically assume a CUDA-capable GPU. We therefore follow a simple recipe suited to low-resource settings: full fine-tuning of a small model, with every cost dimension reported.

\paragraph{Civil-registration digitization in Africa.}
Continental and national programmes are modernizing civil registration~\cite{au2019,uneca2024}, for example Senegal's NEKKAL~\cite{nekkal2023}. These efforts focus mainly on computerizing new registrations and indexing scanned documents at the record level, not on transcribing the content of historical handwritten registers. Our work addresses this gap with an approach sized for local administrations rather than national programmes.

% =====================================================================
\section{Dataset and Capture Protocol}
\label{sec:data}

\paragraph{Source and context.}
The corpus comes from the civil-status archive of a municipal registry in Chad. Chad's recent civil-registration reforms focus on secure identity documents, late registration of children, and a new biometric register~\cite{tchadinfos2026,realites2025}. None of them addresses the existing stock of handwritten registers, which remains the only record of many past births, marriages, and deaths. The registry we worked with lies outside the urban areas targeted by these programmes and is representative of the many local offices where paper registers are the sole copy.

The registers were kept by a succession of civil-status officers, each with their own handwriting and no common calligraphic standard. This makes the writer population far more varied than in many reference HTR datasets. Most records are pre-printed forms, several of them bilingual French--Arabic, whose handwritten fields are filled in French. The handwritten fields are dominated by local personal and place names that rarely appear in European training corpora. The registers also show physical wear typical of such archives, including stains, tears, faded ink, and damage from humidity and handling.

\paragraph{Capture protocol.}
Collection took place on site over five consecutive days, producing 1,568 images, an average of about 314 per day. This throughput requirement shaped the protocol, which uses only consumer hardware: a Google Pixel~6a smartphone capturing in RAW (DNG) format at an image resolution of $2268\times4032$ pixels; a stand holding the phone parallel to the register, whose height was adjusted for each record; and a ThinkPad T460 laptop running scrcpy (v4.1), which mirrors the phone screen for real-time framing control and remote triggering. We captured in RAW rather than JPEG to avoid the phone's built-in sharpening, denoising, and tone mapping, which can destroy the fine intensity variations our cropping pipeline relies on. Holding the phone parallel to the page and triggering remotely reduce motion blur and perspective distortion, which is therefore limited at capture time rather than corrected in software. Lighting was not controlled: captures were made under ambient conditions, so the page-detection stage must cope with uneven illumination. The full setup (laptop, phone, and capture stand) costs about 150,000~FCFA (roughly 230~EUR) and requires no specialized scanning equipment.

\paragraph{Corpus composition.}
Table~\ref{tab:corpus} gives the distribution of images by record type. Birth records dominate, and marriage records are scarce. The legal record type is only a rough guide to layout: manual inspection revealed five distinct templates, namely (G1) narrative marriage record, (G2) birth record by direct declaration, (G3) birth record transcribing a supplementary court judgment, (G4) narrative death record, and (G5) birth declaration on a pre-printed form filled in by hand. Birth records alone span three structurally different templates, so the recognizer must handle different line lengths, field arrangements, and mixes of printed and handwritten text.

\begin{table}[t]
\centering
\begin{tabular}{lrr}
\hline
Record type & Images & Share \\
\hline
Birth    & 1,101 & 70.2\% \\
Marriage &   152 &  9.7\% \\
Death    &   315 & 20.1\% \\
\hline
Total    & 1,568 & 100\% \\
\hline
\end{tabular}
\caption{Corpus composition by record type (number of images).}
\label{tab:corpus}
\end{table}

\paragraph{Annotation.}
We transcribed 185 records covering all three record types in eScriptorium~\cite{kiessling2019escriptorium}, producing line-level transcriptions aligned with the image. Lines were first segmented automatically and then corrected manually once; this corrected segmentation was stored and reused every time it was needed (training, evaluation, benchmarking). A line is transcribed if and only if it contains handwritten text: printed text is part of the ground truth only when it appears on a line that also contains handwriting, and purely printed lines are excluded. The data were split 70/15/15 into training, validation, and test sets at the line level, with the constraint that lines from the same record never appear in more than one set. For the downstream task, a separate set of 179 records was annotated with eight entity types in Label Studio (Section~\ref{sec:downstream}). All annotation was carried out by a single annotator, so we cannot report inter-annotator agreement.

\paragraph{Data availability.}
The records contain personal information about real individuals. Only two isolated text lines, with personal names and record identifiers masked, are shown in this paper (Figure~\ref{fig:qual}); the corpus (images, transcriptions, and annotations) is not publicly released (Section~\ref{sec:ethics}). We will release the preprocessing code; the recognizer weights will be released after a memorization check and the required authorization.

% =====================================================================
\section{Method}
\label{sec:method}

Our pipeline has three stages (Figure~\ref{fig:pipeline}): classical \emph{preprocessing} that turns raw smartphone captures into cropped, correctly oriented images of single records; \emph{handwritten text recognition} with a fine-tuned Kraken model; and a \emph{downstream extraction} stage that we use only to measure how recognition errors propagate. Inference in every stage runs on a laptop CPU.

\begin{figure}[t]
\centering
\resizebox{\linewidth}{!}{% figs/pipeline.tex -- pipeline overview (replaces the Figure 1 placeholder)
% Requires in preamble: \usepackage{tikz}
%   \usetikzlibrary{positioning,arrows.meta,fit,backgrounds,calc}
% Solid fill  = learned component; dashed = fixed rules / classical ops.
\begin{tikzpicture}[
  font=\footnotesize,
  node distance=4mm and 5mm,
  >={Stealth[length=2mm]},
  stage/.style={draw, rounded corners=2pt, align=center, minimum height=11mm,
                text width=19mm, inner sep=2pt},
  learned/.style={stage, fill=blue!12, draw=blue!60!black, thick},
  rules/.style={stage, fill=gray!8, draw=gray!60!black, dashed, thick},
  io/.style={stage, fill=white, draw=black!70},
  lbl/.style={font=\scriptsize\itshape, text=black!70},
  flow/.style={->, thick, black!75}
]

% ---------------- Row 1: vision pipeline (paper's focus) ----------------
\node[io]      (cap)  {Smartphone\\RAW capture\\{\scriptsize fixed stand}};
\node[rules, right=of cap] (pre)
      {Classical\\preprocessing\\{\scriptsize no learned model}};
\node[learned, right=of pre] (htr)
      {Kraken HTR\\CRNN-CTC\\{\scriptsize fine-tuned}};
%\node[rules, right=of htr, text width=17mm] (ptq)
      %{INT8 PTQ\\{\scriptsize post-training}};
\node[io, right=of htr, text width=16mm] (txt) {Line\\transcriptions};

\draw[flow] (cap) -- (pre);
\draw[flow] (pre) -- (htr);
\draw[flow] (htr) -- (txt);
%\draw[flow] (ptq) -- (txt);

% ---------------- Row 2: downstream evaluation probe ----------------
\node[learned, below=10mm of htr] (ner)
      {NER\\{\scriptsize camembert-ner-}\\[-1pt]{\scriptsize with-dates}};
\node[rules, left=of ner] (tpl)
      {Template\\classification\\{\scriptsize G1--G5}};
\node[rules, left=of tpl] (role)
      {Role\\resolution\\{\scriptsize $O(n)$ per record}};
\node[io, right=of ner, text width=17mm] (gt) {Ground-truth\\text\\{\scriptsize (upper bound)}};

\draw[flow] (txt.south) |- ([yshift=4mm]ner.north) -- (ner.north);
\draw[flow, dashed] (gt) -- (ner);
\draw[flow] (ner) -- (tpl);
\draw[flow] (tpl) -- (role);

% ---------------- Group frames ----------------
\begin{scope}[on background layer]
  \node[draw=blue!50!black, fill=blue!3, rounded corners=4pt, inner sep=3mm,
        fit=(cap)(pre)(htr)(txt)] (g1) {}; %(ptq)
  \node[draw=orange!70!black, fill=orange!4, rounded corners=4pt, inner sep=3mm,
        fit=(role)(tpl)(ner)(gt)] (g2) {};
\end{scope}
\node[lbl, anchor=south west] at (g1.north west)
     {Vision pipeline (Secs.~4.1--4.3) \;\textbullet\; CPU-only inference};
\node[lbl, anchor=north west] at (g2.south west)
     {Downstream evaluation probe (Sec.~4.4) \;\textbullet\; error propagation, Sec.~5.4};

% ---------------- Legend ----------------
\node[learned, text width=6mm, minimum height=3mm, inner sep=0pt,
      below=7mm of g2.south east, anchor=north east, xshift=-24mm] (lk1) {};
\node[right=1mm of lk1, font=\scriptsize] {learned};
\node[rules, text width=6mm, minimum height=3mm, inner sep=0pt,
      right=13mm of lk1] (lk2) {};
\node[right=1mm of lk2, font=\scriptsize] {rules / classical};
\end{tikzpicture}}
\caption{Overview of the pipeline, with CPU-only inference. Top: the vision pipeline evaluated in this paper. Bottom: the downstream extraction stages, used only as a probe to measure how recognition errors propagate; feeding them ground-truth text gives the upper bound in Table~\ref{tab:prop}.}
\label{fig:pipeline}
\end{figure}

\subsection{Classical preprocessing}
\label{sec:prepro}
We deliberately use no learned model in preprocessing: classical operations need no training data or annotation, have predictable failure modes, and run fast on commodity hardware.

\paragraph{Normalization and orientation.}
Raw DNG files are renamed deterministically using their MD5 hash and grouped into batches. A portrait/landscape check found no image that needed rotation in practice. Upside-down captures occurred in a subset of the images of the marriage sub-corpus: 87 images in total, or 5.5\% of the corpus. We detect them by normalized cross-correlation template matching of the letter ``e'', a frequent character in the corpus, with a detection threshold of 0.80, and rotate the detected images by 180\textdegree{}. Finally, we remove a fixed strip along the right edge that contains no document content.

\paragraph{Page cropping.}
Smartphone captures taken under uncontrolled lighting have uneven illumination, so a single global threshold such as Otsu's~\cite{otsu1979} is unreliable. Instead, after smoothing with a $7\times7$ Gaussian kernel, we combine adaptive Gaussian thresholding with an HSV criterion that keeps low-saturation, bright pixels ($S<70$ and $V>110$ on an 8-bit scale) to separate page from background. We then compute a Canny edge map~\cite{canny1986} (thresholds 30 and 90), reinforce it with directional morphological closing (vertical $3\times60$ and horizontal $60\times3$ kernels), and take horizontal and vertical projection profiles of the result to locate the page boundaries.

\paragraph{Separating adjacent records.}
When the detected region is too wide for a single record, because two records were captured in one frame, a disambiguation step chooses the split that best separates them, scoring candidates on fill ratio, relative width, and edge strength. A second pass targeted at the left margin then removes the binding area: after a $21\times21$ Gaussian blur and contrast-limited adaptive histogram equalization (clip limit 3.0), a Canny edge map (thresholds 100 and 200) locates the binding, and a fixed offset of 45 pixels is applied to the left boundary.

\paragraph{No perspective correction.}
We do not apply homography-based rectification. The capture protocol keeps the phone parallel to the page, so perspective distortion is limited at the source: we prevent distortion during capture rather than correct it with extra computation afterwards. Some residual curvature near the binding remains (Section~\ref{sec:disc}).

The full pipeline ran on all 1,568 images without errors, at about 2.6 images per second on the Dell~7750 laptop used for the benchmarks (Section~\ref{sec:results}). We did not measure the accuracy of page crops, record splits, or upside-down detection against manual references; the parameters (Table~\ref{tab:prepro}) were tuned by hand on this corpus and may need retuning elsewhere.

\begin{table}[t]
\centering
\small
\begin{tabular}{lll}
\hline
Stage & Parameter (code name) & Value \\
\hline
Page detection        & \texttt{blur\_ksize}                              & (7,7) \\
                      & \texttt{s\_max}, \texttt{v\_min}                  & 70, 110 \\
                      & \texttt{canny\_t1}, \texttt{canny\_t2}            & 30, 90 \\
                      & \texttt{close\_v}, \texttt{close\_h}              & (3,60), (60,3) \\
                      & \texttt{prof\_thr}                                & 0.5 \\
                      & \texttt{ratio\_min\_page}, \texttt{ratio\_cible}  & 1.9, 2.35 \\
                      & \texttt{snap\_tol}, \texttt{margin}               & 45, 10 \\
\hline
Left-margin cleaning  & \texttt{blur\_ksize\_s2}                          & (21,21) \\
                      & \texttt{clahe\_clip\_s2}                          & 3.0 \\
                      & \texttt{canny\_t1\_s2}, \texttt{canny\_t2\_s2}    & 100, 200 \\
                      & \texttt{decalage\_gauche\_px}                     & 45 \\
\hline
Upside-down detection & template character, NCC threshold                 & ``e'', 0.80 \\
\hline
\end{tabular}
\caption{Preprocessing parameters (code names as in our implementation).}
\label{tab:prepro}
\end{table}

\subsection{Handwritten text recognition}
\label{sec:htr}

\paragraph{Architecture.}
We use Kraken~\cite{kiessling2019kraken}, an open-source HTR system built on a hybrid convolutional-recurrent network trained with the CTC loss~\cite{graves2006ctc}. It is small enough to run on a CPU, can be fine-tuned on a free-tier GPU, integrates directly with the eScriptorium annotation platform, and is widely used for historical documents.

\paragraph{Line segmentation.}
Lines were segmented automatically and then corrected manually once (Section~\ref{sec:data}); this corrected segmentation is reused throughout. All reported CER and WER values are therefore computed on known, ground-truth line boundaries and do not include segmentation errors. End-to-end evaluation with automatic segmentation is left to future work.

\paragraph{Fine-tuning.}
Training from scratch is infeasible with a few hundred annotated records. We therefore fully fine-tune all weights of the pretrained Kraken model \texttt{kraken-ppocrv6-medium} (Hugging Face: \texttt{small-models-for-glam/kraken-ppocrv6-medium}). Because the pretrained output alphabet does not cover every character and diacritic in our corpus, we merge it with the alphabet of our data using Kraken's \texttt{-{}-resize union} option. We train for 40 epochs with a learning rate of $10^{-3}$ and a batch size of 7. Training took about four hours on a free-tier Google Colab GPU; it is the only stage of the pipeline that used a GPU. We did not use parameter-efficient methods: the model is small enough that full fine-tuning is already inexpensive.

\paragraph{Data split.}
Ground truth is split 70/15/15 at the line level, Kraken's native unit, with the constraint that lines from one record never appear in more than one split. This prevents record-level leakage but not writer-level leakage (Section~\ref{sec:disc}).

\subsection{Cost measurement protocol}
\label{sec:frugal}

We report quality jointly with cost. We compare two configurations that share the same architecture and FP32 precision: the pretrained model and the fine-tuned model. Fine-tuning changes the weights but not the network, so any difference in inference cost between them reflects measurement noise rather than model size. We assume an operational ceiling of 2\% CER, above which we expect checking and correcting the output to approach the effort of partial manual transcription. This threshold is a working assumption, not a measured correction cost, and the downstream results (Section~\ref{sec:res-prop}) suggest that the error rate on names matters more than overall CER.

\paragraph{Benchmark protocol.}
We process 50 images drawn at random (seed 42) in five steps of ten images. Both models process the same images, and two additional images are processed beforehand as warm-up and excluded. Inference uses batches of 8 lines, line-length bucketing (64) with sorting, a frozen computation graph, channels-last layout, batch-normalization fusion, 6 threads, 2 line workers, and a prefetch of 2. The measured time covers recognition on the stored, manually corrected line segmentation and excludes automatic line segmentation.

\paragraph{Metrics.}
We report weight file size, model loading time, one-time warm-up time (first inference after loading), peak resident memory (RSS) of the process, time per image and per line, throughput, and mean CPU utilization. Time per image is the mean over the five steps, with its standard deviation across steps. We also report the training cost of fine-tuning.

We did not evaluate reduced-precision inference (e.g.\ INT8 quantization). Its benefit depends on hardware support and on the inference runtime, and our target machines are ordinary laptop CPUs without dedicated low-precision instructions; we leave it to future work on the actual deployment hardware.

\subsection{Downstream extraction (evaluation probe)}
\label{sec:downstream}
To measure how recognition quality affects a real task, we feed HTR output into an extraction stage that identifies the people in each record and their roles (father, mother, spouse, witness, etc.). We fully fine-tune camembert-ner-with-dates~\cite{camembertnerdates}, a French model of roughly 110M parameters derived from CamemBERT~\cite{martin2020camembert}, on 179 annotated records, with eight entity types: PER, LOC, ORG, and DATE (inherited from the base model) and PROFESSION, NUM\_REF (reference numbers), ACTE\_TYPE (record type), and ANCRE (lexical role markers such as \emph{fils de}, ``son of''). The record's template is inferred with simple rules from the detected ACTE\_TYPE entities. Roles are then assigned in linear time by a deterministic algorithm that walks through the entities in reading order and uses a template-specific dictionary mapping each marker to a role; each person entity inherits the role set by the most recent marker, and a block-counting mechanism handles the repeated field groups of template G5. Because these rules are fixed, role errors come from only three sources, namely recognition, NER, or template classification, which lets us isolate the contribution of HTR (Section~\ref{sec:res-prop}).

% =====================================================================
\section{Experiments and Results}
\label{sec:results}

\subsection{Setup}
Preprocessing and inference benchmarks run on a Dell~7750 mobile workstation with an Intel Core i7-10850H CPU (6 cores), 15.6\,GiB of RAM, and no GPU. This machine is faster than the ThinkPad T460 used for capture, on which we did not run the benchmarks. Fine-tuning ran on a free-tier Google Colab GPU. Software: Python 3.13, PyTorch 2.14.0 (CPU build), Kraken 7.1.1, OpenCV 5.0.0.93, jiwer 4.0.0, seqeval 1.2.2. Character error rate is
\begin{equation}
\mathrm{CER}=\frac{S+D+I}{N},
\end{equation}
where $S$, $D$, $I$ are character substitutions, deletions, and insertions and $N$ is the number of reference characters; WER is defined analogously at word level.

\subsection{Recognition quality}
\label{sec:res-quality}

\begin{table}[t]
\centering
\begin{tabular}{lcc}
\hline
Model & CER (\%) $\downarrow$ & WER (\%) $\downarrow$ \\
\hline
Kraken, pretrained (no adaptation) & 14.53 & 56.93 \\
Kraken, fine-tuned                 & 0.50  & 3.37 \\
\hline
\end{tabular}
\caption{HTR results on the test set, computed on ground-truth line segmentation.}
\label{tab:htr}
\end{table}

\begin{figure}[t]
\centering
\includegraphics[width=\linewidth]{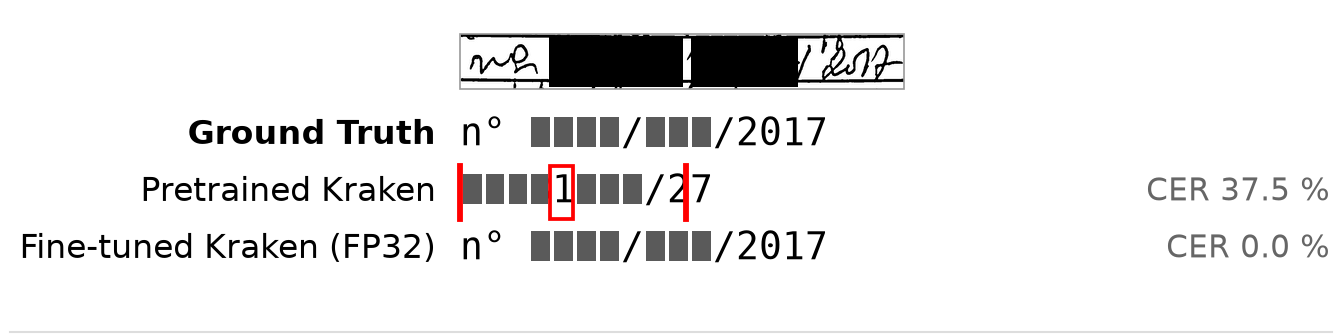}\\[4pt]
\includegraphics[width=\linewidth]{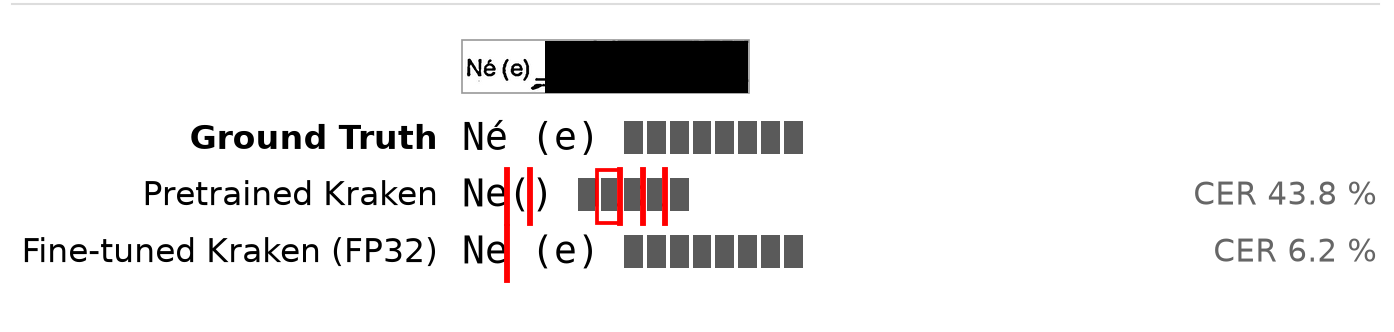}
\caption{Two example test lines. Personal names, the record number and the office code are masked in the images (black) and replaced character by character in the transcriptions (grey); red marks show recognition errors.}
\label{fig:qual}
\end{figure}

Fine-tuning reduces CER by a factor of about 29 and WER by a factor of about 17 (Table~\ref{tab:htr}). In absolute terms WER falls by 53.6 points, against 14.0 points for CER: with the pretrained model more than half of the words are affected by errors, whereas after fine-tuning fewer than 4\% are. On inspection, many of the pretrained model's errors fall on local personal and place names, as reported for other civil records~\cite{hoek2026haystack}, but we did not measure error rates on name spans. Part of its 14.53\% CER also comes from characters absent from its output alphabet (Section~\ref{sec:htr}), which it cannot produce at all. Both error rates are computed on the manually corrected line segmentation, so they measure recognition alone, not segmentation followed by recognition. Figure~\ref{fig:qual} shows representative test lines. Two further factors make the 0.50\% CER optimistic: test lines are written by officers who also wrote training lines, and the ground truth includes printed field labels (\eg ``N\'e (e)'') whenever they share a line with handwriting; these labels recur across records and are easy to learn. We therefore read the figure as evidence of feasibility within this registry rather than as an estimate of performance on unseen writers or other registries (Section~\ref{sec:disc}).

\subsection{Frugality of the recognizer}
\label{sec:res-frugal}

\begin{table}[t]
\centering
\resizebox{\linewidth}{!}{%
\begin{tabular}{lcccccc}
\hline
Model & Disk & Load / warm-up & Peak RSS & Time / image & Time / line & Images / min \\
      & (MiB) & (s) & (GiB) & (s) & (ms) & \\
\hline
Pretrained & 60.8 & 1.1 / 26.1 & 2.30 & $5.84 \pm 1.13$ & 282 & 10.3 \\
Fine-tuned & 60.8 & 1.1 / 25.3 & 2.49 & $5.60 \pm 1.09$ & 270 & 10.7 \\
\hline
\end{tabular}}
\caption{Inference cost on a six-core laptop CPU (Intel Core i7-10850H, no GPU), for 50 images in five steps of 10 images (1,037 lines). Peak RSS is the mean over steps; $\pm$ is the standard deviation across steps.}
\label{tab:frugal}
\end{table}

As expected for two models with the same architecture, their inference costs are similar (Table~\ref{tab:frugal}); the table mainly describes what one Kraken recognizer costs on a laptop CPU. Both models have the same weight file (60.8\,MiB) and load in about one second. Peak memory stays below 2.6\,GiB for both, far under the 15.6\,GiB available; we did not test smaller machines. The small gaps between the two models (about 4\% in time, about 0.2\,GiB in peak RSS) are of the same order as the step-to-step variation and may depend on execution order, so we do not interpret them. The first inference after loading takes about 25\,s and is a one-time cost paid at startup.

Throughput is about 10 images per minute, or 3.5--3.7 lines per second, using on average 5.3 of the 6 cores. Extrapolated to the full corpus, recognition of the 1,568 images takes roughly 2.5 hours on a single laptop. The first two steps were slower for both models (95th-percentile latency around 15\,s, about 4.5 busy cores) than the last three (about 4.8--5.1\,s per image, 5.8 busy cores); since both models process the same images, we attribute this to system effects such as thermal throttling or cache state rather than to image content.

\paragraph{Training cost.}
Fine-tuning (40 epochs) took about four hours on a free-tier Google Colab GPU. This is a one-off cost, but it is the only step that required a GPU; we did not measure CPU-only training.

Since the fine-tuned model reaches 0.50\% CER on this test set, below our assumed 2\% ceiling, and runs as a small recurrent-convolutional model on a laptop CPU, it meets our working criteria on the dimensions we measured, subject to the caveats of Section~\ref{sec:res-quality}.

\subsection{How recognition errors propagate}
\label{sec:res-prop}

\begin{table}[t]
\centering
\begin{tabular}{lc}
\hline
Input text & Role-resolution F1 (\%) \\
\hline
Ground-truth transcription (upper bound) & 90.3 \\
Fine-tuned Kraken                        & 81.2 \\
Pretrained Kraken (no adaptation)        & 44.5 \\
\hline
\end{tabular}
\caption{Role-resolution F1 as a function of input text quality ($n=46$ records, none from the HTR test set).}
\label{tab:prop}
\end{table}

We keep the extraction stage fixed and vary only its input text, on a subsample of 46 records that are not part of the HTR test set (Table~\ref{tab:prop}). Because these records are outside the test set, the fine-tuned recognizer may have seen some of them during training, so its score is best read as an optimistic estimate and the recognition cost as a lower bound. Three observations follow. First, the extraction stage has its own ceiling: even with perfect text, role resolution reaches only 90.3\% F1, owing to NER and template-classification errors. Second, residual recognition errors cost about 9 F1 points despite a test-set CER of only 0.50\%: we expect recognition errors to concentrate on names and role markers, exactly the tokens the downstream task depends on. Third, with the pretrained recognizer F1 drops by a further 36.7 points, making the task essentially unusable. For information extraction, \emph{where} recognition errors fall matters as much as how many there are.

\subsection{Downstream extraction (summary)}
The fine-tuned NER model reaches 98.0\% micro-F1 over the eight entity types, evaluated on ground-truth transcriptions of a held-out test split (1,242 entities; the 179 records are split 70/15/15 by record). On the three entity types shared with the base model (PER, LOC, and DATE; ORG is left out because it is too rare in our data to be informative), adaptation improves weighted F1 from 86.8\% to 98.6\%, with most of the gain on PER entities (+18.0 points, against +8.0 for DATE). This suggests that the base NER model is weakest on local personal names. Rule-based template classification reaches 87.0\% accuracy ($n=100$); 8 of its 13 errors confuse templates G2 and G3, which share the same role sequence, so this confusion does not harm role resolution.

% =====================================================================
\section{Discussion and Limitations}
\label{sec:disc}

\paragraph{What the CER gain does and does not show.}
The 29-fold CER reduction shows that adaptation helps on this registry, but it does not explain why. At least three factors are confounded: the pretrained model is a general-purpose recognizer rather than a French handwriting model; it cannot output some characters of our corpus; and the fine-tuned model is tested on writers it has already seen. Our observation that errors fall on names is qualitative, and the stronger NER gain on PER than on DATE (+18.0 vs. +8.0 F1) concerns a different model. That generic models are poorly suited to local names is therefore a hypothesis to test, with name-span CER, an out-of-alphabet breakdown and a handwriting-specific baseline, rather than a result of this paper.

\paragraph{Frugality is not a single number.}
Fine-tuning leaves the architecture, and hence the inference cost, unchanged by design. The costs that matter for deployment lie elsewhere: a one-time warm-up of about 25\,s, a peak memory of about 2.5\,GiB, a throughput of about 10 images per minute, and, for adaptation itself, about four hours of GPU training plus the human effort of transcribing 185 records, correcting their segmentation and annotating 179 records for NER, which we did not time. A record-holding administration can run the resulting model on a laptop CPU, but adapting it to a new registry currently requires access to a GPU, or CPU training that we did not evaluate. Frugality claims should report every cost dimension, and we did not test lower-end machines or reduced-precision inference.

\paragraph{Limitations.}
\emph{Single site and possible writer overlap.} All records come from one registry. The line-level split keeps each record within one split, but lines by the same officer, or from the same register, can appear in both training and test data, so the 0.50\% CER may overestimate performance on unseen writers or other registries. A split by register or by writer, and a second site, are the most important next steps.
\emph{Small test set.} The HTR test set contains about 28 records, and marriage and death records are under-represented; per-type results are diagnostic at best, and we report no confidence intervals.
\emph{What the CER measures.} Printed field labels are part of the ground truth when they share a line with handwriting, which lowers CER; we do not report CER on handwritten characters only, on names, or per template. The recognizer was trained once, with no repeated runs or variance estimate, and compared with a single, general-purpose baseline.
\emph{Single annotator.} We cannot estimate how much residual error reflects annotation ambiguity.
\emph{Line segmentation.} CER and WER assume the manually corrected line segmentation; errors of automatic segmentation are not measured, and the timings exclude it.
\emph{Cost measurements.} Timings come from 50 images on a single six-core laptop, with the two models run sequentially, and show a systematic slowdown in the first two steps; they are orders of magnitude, not precise figures. Older machines, such as the laptop used for capture, will likely be slower. Quality and cost were measured on different image samples, and the cost-benchmark images were drawn from the corpus and may overlap with the annotated records.
\emph{Preprocessing coverage.} Upside-down detection was developed and tested only on the marriage sub-corpus, where such captures occurred, and no perspective correction is applied near the binding.
\emph{Downstream probe.} The 46 probe records may overlap the training data of the recognizer and of the NER model, so its scores are optimistic.
\emph{Unablated choices.} The benefit of RAW over JPEG capture was not tested.
\emph{Frozen downstream rules.} Role resolution relies on rules tied to this registry's five templates and will likely need rewriting elsewhere.

% =====================================================================
\section{Ethics Statement}
\label{sec:ethics}

The corpus consists of civil-status records of real people, containing names, family relationships, dates and places of birth, marriage, and death, and in some cases national identity numbers. Most of the people concerned were not informed that their records would be used for research. We did not anonymize or pseudonymize the working corpus: models were trained and evaluated on the original, identifying text. To limit risk, we (i)~processed the data locally on a single machine, except for fine-tuning, which ran on a free-tier cloud GPU and therefore involved uploading the training images and transcriptions to that service, (ii)~show no full record in this paper, only two isolated text lines in which personal names and record identifiers are masked (Figure~\ref{fig:qual}), and (iii)~do not release the images, transcriptions, or annotations. The cloud upload runs counter to the aim of this work, which is to keep identity data under the control of the administration and the community that hold it~\cite{fendji2025communitydata}; future adaptation should use hardware controlled by the institution. 

Chad's data-protection framework requires processing of personal data to be declared to, or authorized by, the national agency ANSICE~\cite{tchad_loi007,tchad_loi006,bassounda2020}. Data collection was carried out with the written permission of the Municipality of Koumra, which authorizes use of the records for academic purposes only, requires personal information in published images to be masked, and excludes any commercial use. We do not claim that this study has been declared to or authorized by ANSICE, nor that research use of public civil-status archives is exempt from individual consent under this framework; we will seek the corresponding authorization before any release of data or models.

Fine-tuning on a cloud GPU required uploading the training images and transcriptions to that service. Future adaptation will run on hardware controlled by the institution, so that identity data remain under the control of the administration and the community that hold them~\cite{fendji2025communitydata}.

Systems that make identity records searchable at scale could also support surveillance or profiling. Local, CPU-only deployment under the control of the record-holding administration reduces this risk compared with outsourced processing, but does not remove it.

% =====================================================================
\section{Conclusion}
\label{sec:concl}

We asked whether a frugal pipeline, built from consumer capture hardware, a few hundred annotated records, and CPU-only inference, can make handwritten civil-status registers from Sub-Saharan Africa machine-readable. On a new corpus of 1,568 images of records from a registry in Chad, fine-tuning a Kraken recognizer for about four hours on a free-tier GPU reduced CER from 14.53\% to 0.50\% on corrected line segmentation and writers seen in training, and the model runs at under 6\,s per image and under 2.6\,GiB of memory on a laptop CPU. A downstream extraction task showed that even this small residual error costs about 9 F1 points. These results suggest that recognition for such archives need not require GPUs or cloud services at deployment time, but they come from a single registry, a small test set with writer overlap, and a single baseline. Future work includes evaluation across registers, writers, and sites; error rates on handwritten text and on names; handwriting-specific and Transformer baselines; a learning curve over annotation size; a second annotator to measure agreement; end-to-end evaluation with automatic line segmentation; and, above all, an anonymization scheme that still permits record linkage, which must be in place before any data or trained models are shared with, or deployed by, the administrations that hold these records.

\section*{Acknowledgements}
We thank the Municipality of Koumra (Mandoul Oriental, Chad) for granting access to the civil-status registers used in this study and authorizing their use for academic research.

\bibliography{refs}

\end{document}